\documentclass{article} % For LaTeX2e
\usepackage{iclr2016_conference,times}
\usepackage{hyperref}
\usepackage{url}
\usepackage{times}
\usepackage{graphicx}
\usepackage{latexsym}
\usepackage{amsmath}
\usepackage{amssymb}
\usepackage{paralist}
\usepackage{booktabs}
\usepackage{comment}

\renewcommand{\vec}{\mathbf}

\title{Grid Long Short-Term Memory}

\author{Nal Kalchbrenner \& Ivo Danihelka \& Alex Graves  \\
Google DeepMind \\
London,  United Kingdom\\
\texttt{\{nalk,danihelka,gravesa\}@google.com} \\
}

\begin{document}

\maketitle

\begin{abstract}

This paper introduces \emph{Grid Long Short-Term Memory}, a network of LSTM cells arranged in a multidimensional grid that can be applied to vectors, sequences or higher dimensional data such as images.
The network differs from existing deep LSTM architectures in that the cells are connected between network layers as well as along the spatiotemporal dimensions of the data.
The network provides a unified way of using LSTM for both deep and sequential computation.
We apply the model to algorithmic tasks such as 15-digit integer addition and sequence memorization, where it is able to significantly outperform the standard LSTM.
We then give results for two empirical tasks. We find that 2D Grid LSTM achieves 1.47 bits per character on the Wikipedia character prediction benchmark, which is state-of-the-art among neural approaches. In addition, we use the Grid LSTM to define a novel two-dimensional translation model, the \emph{Reencoder}, and show that it outperforms a phrase-based reference system on a Chinese-to-English translation task.

\end{abstract}

\section{Introduction}

Long Short-Term Memory (LSTM) networks are recurrent neural networks equipped with a special gating mechanism that controls access to memory cells~\citep{hochreiter97lstm}. Since the gates can prevent the rest of the network from modifying the contents of the memory cells for multiple time steps, LSTM networks  preserve signals and propagate errors for much longer than ordinary recurrent neural networks. By independently reading, writing and erasing content from the memory cells, 
the gates can also learn to attend to specific parts of the input signals and ignore other parts.
These properties allow LSTM networks to process data with complex and separated interdependencies and to excel in a range of sequence learning domains such as speech recognition \citep{graves13drnn}, offline hand-writing recognition \citep{graves09nips}, machine translation \citep{sutskever2014sequence} and image-to-caption generation \citep{vinyals2014show,DBLP:journals/corr/KirosSZ14}.

Even for non-sequential data, the recent success of deep networks has shown
that long chains of sequential computation are key to finding and exploiting complex patterns. Deep networks suffer from exactly the same problems as recurrent networks applied to long sequences: namely that information from past computations rapidly attenuates as it progresses through
the chain -- the \emph{vanishing gradient problem} \citep{Hochreiter:91} -- and that each layer cannot dynamically select or ignore its inputs.
 It therefore seems attractive to generalise the advantages of LSTM to deep computation.

We extend LSTM cells to deep networks within a unified architecture. We introduce Grid LSTM, a network that is arranged in a grid of one or more dimensions. The network has LSTM cells along any or all of the dimensions of the grid. The depth dimension is treated like the other dimensions and also uses LSTM cells to communicate directly from one layer to the next.
Since the number $N$ of dimensions in the grid can easily be 2 or more, we propose a novel, robust way for modulating the \emph{N}-way communication across the LSTM cells.

\emph{N}-dimensional Grid LSTM (\emph{N}-LSTM for short) can naturally be applied as feed-forward networks as well as recurrent ones. One-dimensional Grid LSTM corresponds to a feed-forward network that uses LSTM cells in place of transfer functions such as $\tanh$ and ReLU \citep{DBLP:conf/icml/NairH10}. These networks are related to Highway Networks \citep{srivastava2015highway} where a gated transfer function is used to successfully train feed-forward networks with up to 900 layers of depth.
Grid LSTM with two dimensions is analogous to the \emph{Stacked LSTM}, but it adds cells along the depth dimension too. Grid LSTM with three or more dimensions is analogous to \emph{Multidimensional LSTM} \citep{graves13drnn,sutskever2014sequence,graves07mdlstm,graves12supervised}, but differs from it not just by having the cells along the depth dimension, but also by using the proposed mechanism for modulating the \emph{N}-way interaction that is not prone to the instability present in Multidimesional LSTM.

We study some of the learning properties of Grid LSTM in various algorithmic tasks. We compare the performance of two-dimensional Grid LSTM to Stacked LSTM on computing the addition of two 15-digit integers without curriculum learning  and on memorizing sequences of numbers \citep{DBLP:journals/corr/ZarembaS14}. We find that in these settings having cells along the depth dimension is more effective than not having them; similarly, tying the weights across the layers is also more effective than untying the weights, despite the reduced number of parameters.  

We also apply Grid LSTM to two empirical tasks. The architecture achieves 1.47 bits-per-character in the 100M characters Wikipedia dataset  \citep{hutterDataset} outperforming other neural networks.  Secondly, we use Grid LSTM to define a novel neural translation model that \emph{re-encodes} the source sentence based on the target words generated up to that point. The network  outperforms the reference phrase-based CDEC system  \citep{cdec} on the IWSLT BTEC Chinese-to-Ensligh translation task. The appendix contains additional results for Grid LSTM on learning parity functions and classifying MNIST images.

The outline of the paper is as follows. In Sect.~2 we describe standard LSTM networks that comprise the background. In Sect.~3 we define the Grid LSTM architecture. In Sect.~4 we consider the six experiments and we conclude in Sect.~5.

\begin{figure}
\centering
\includegraphics[width=\textwidth]{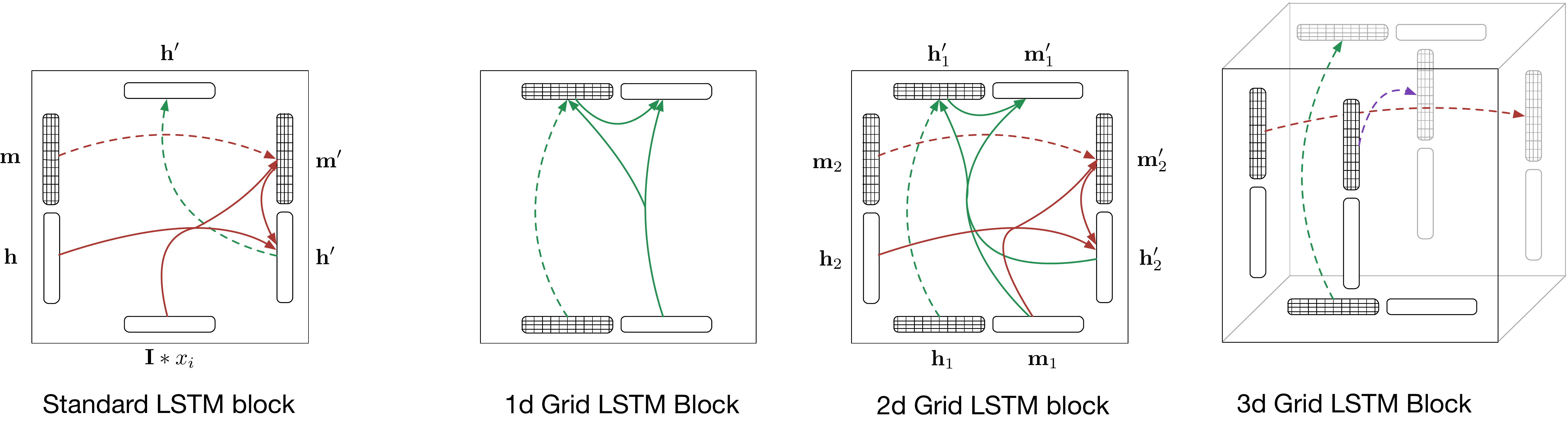}
\caption{Blocks form the standard LSTM  and those that form Grid LSTM networks of $N=1,2$ and $3$ dimensions. The dashed lines indicate identity transformations. The standard LSTM block does not have a memory vector in the vertical dimension; by contrast, the  2d Grid LSTM block has the memory vector $\vec{m}_1$ applied along the vertical dimension.}
\end{figure}

\section{Background}

We begin by describing the standard LSTM recurrent neural network and the derived \emph{Stacked} and \emph{Multidimensional} LSTM networks; some aspects of the networks motivate the Grid LSTM.

\subsection{Long Short-Term Memory}

The LSTM network processes a sequence of input and target pairs $(x_1,y_1),...,(x_m,y_m)$. For each pair $(x_i,y_i)$ the LSTM network takes the new input $x_i$ and produces an estimate for the target $y_i$ given all the previous inputs $x_1,...,x_i$. The past inputs $x_1,...,x_{i-1}$ determine  the {state} of the network that comprises a \emph{hidden} vector $\vec{h}\in\mathbb{R}^d$ and a \emph{memory} vector $\vec{m}\in\mathbb{R}^d$.  The computation at each step is defined as follows \citep{graves13drnn}:
\begin{align}
\label{lstmtrans}
\begin{split}
\vec{g}^{u} &= \sigma(\vec{W}^u \vec{H}) \\		
\vec{g}^f  &= \sigma(\vec{W}^f  \vec{H}) \\
\vec{g}^o &= \sigma(\vec{W}^o  \vec{H})\\
\vec{g}^c &= \tanh(\vec{W}^c \vec{H}) \\
\vec{m}' &= \vec{g}^f \odot \vec{m} + \vec{g}^u \odot \vec{g}^c  \\
\vec{h}' &= \tanh(\vec{g}^o \odot \vec{m}')  \\
\end{split}
\end{align}
where $\sigma$ is the logistic sigmoid function, $\vec{W}^u,\vec{W}^f,\vec{W}^o,\vec{W}^c$  in $\mathbb{R}^{d \times 2d}$ are the recurrent weight matrices of the network and $\vec{H} \in \mathbb{R}^{2d}$ is the concatenation of the new input $x_i$, transformed by a projection matrix $I$, and the previous hidden vector $\vec{h}$:
\begin{equation}
\label{hh}
\vec{H} = \begin{bmatrix}
    I  {x}_{i}    \\
    \vec{h}  \\
\end{bmatrix}
\end{equation}
The computation outputs new hidden and memory vectors $\vec{h}'$ and $\vec{m}'$ that comprise the next state of the network. The estimate for the target  is then computed in terms of the hidden vector $\vec{h}'$. We use the functional $\mathsf{LSTM}(\cdot, \cdot, \cdot)$ as shorthand for Eq.~1 as follows:
\begin{equation}
(\vec{h}', \vec{m}') = \mathsf{LSTM}(\vec{H},\vec{m}, \vec{W})
\end{equation}
where $\vec{W}$ concatenates the four weight matrices $\vec{W}^u, \vec{W}^f,\vec{W}^o,\vec{W}^c$.

One aspect of LSTM networks is the role of the gates $\vec{g}^u,\vec{g}^f,\vec{g}^o$ and $\vec{g}^c$. The forget gate $\vec{g}^f$ can delete parts of the previous memory vector $\vec{m}_{i-1}$ whereas the gate $\vec{g}^c$ can write new content to the new memory $\vec{m}_i$ modulated by the input gate $\vec{g}^u$. The output gate controls what is then read from the new memory $\vec{m}_i$ onto the hidden vector $\vec{h}_i$. The mechanism has two important learning properties. Each memory vector is obtained by a \emph{linear} transformation of the previous memory vector and the gates; this ensures that the forward signals from one step to the other are not repeatedly squashed by a non-linearity such as $\tanh$ and that the backward error signals do not decay sharply at each step, an issue known as the vanishing gradient problem  \citep{Hochreiter2001}.  The mechanism also acts as a memory and {implicit attention} system, whereby the signal from some input $x_i$ can be written to the memory vector and attended to in parts across multiple steps by being retrieved one part at a time. 

\begin{figure}
\centering
\includegraphics[width=\textwidth]{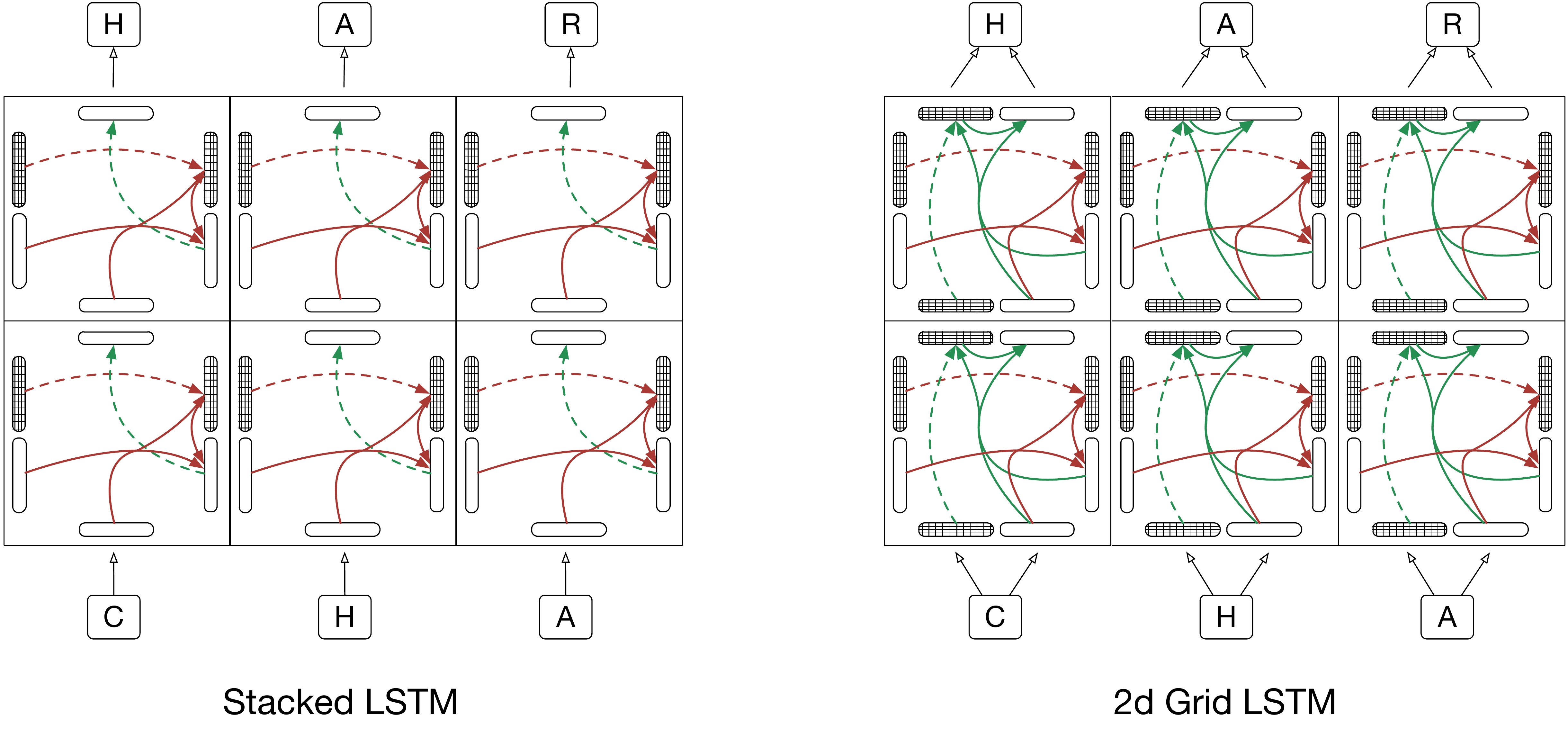}
\caption{Stacked LSTM and 2d Grid LSTM applied to character prediction composed from the respective blocks (Fig.~1). Note how in the Grid LSTM the signal flows through LSTM cells (shaded rectangles) along both the time and the depth dimensions.}
\end{figure}

\subsection{Stacked LSTM}

A model that is closely related to the standard LSTM network is Stacked LSTM \citep{graves13drnn,sutskever2014sequence}. Stacked LSTM adds capacity by stacking LSTM layers on top of each other. The output hidden vector $\vec{h}_i$ in Eq.~1 from the LSTM below is taken as the input to the LSTM above in place of  $I * {x}_{i} $. The Stacked LSTM is depicted in Fig.~2. Note that although the LSTM cells are present along the sequential computation of each LSTM network, they are not present in the vertical computation from one layer to the next. 
 
\subsection{Multidimensional LSTM}

Another related model is  Multidimensional LSTM \citep{graves07mdlstm}. Here the inputs are not arranged in a sequence, but in a $N$-dimensional grid, such as the two-dimensional grid of pixels in an image. At each input $x$ in the array the network receives $N$ hidden vectors $\vec{h}_1,...,\vec{h}_N$ and $N$ memory vectors $\vec{m}_1,...,\vec{m}_N$ and computes a hidden vector $\vec{h}$ and a memory vector $\vec{m}$ that are passed as the next state for each of the $N$ dimensions. The network concatenates the transformed input $\vec{I}*{x}$ and the $N$ hidden vectors $\vec{h}_1,...,\vec{h}_N$ into a vector $\vec{H}$ and as in Eq.~1 computes $\vec{g}^u,\vec{g}^o$ and $\vec{g}^c$, as well as $N$ forget gates $\vec{g}_i^f$. These gates are then used to compute the memory vector as follows:
\begin{align}
\label{lstmtrans}
\begin{split}
\vec{m} &= \sum_i^N \vec{g}_i^f \odot \vec{m}_i + \vec{g}^u \odot \vec{g}^c  \\
\end{split}
\end{align}
As the number of paths in a grid grows combinatorially with the size of each dimension and the total number of dimensions $N$, the values in $\vec{m}$ can grow at the same rate due to the unconstrained summation in Eq.~4. This can cause instability for large grids, and adding cells along the depth dimension increases $N$  and exacerbates the problem. This motivates the simple alternate way of computing the output memory vectors in the Grid LSTM. 

\section{Architecture}

Grid LSTM deploys cells along any or all of the dimensions including the depth of the network. In the context of predicting a sequence, the Grid LSTM has cells along two dimensions, the temporal one of the sequence itself and the vertical one along the depth. To modulate the interaction of the cells in the two dimensions, the Grid LSTM 
proposes a simple mechanism where the values in the cells cannot grow combinatorially as in Eq.~4. In this section we describe the multidimensional \emph{blocks} and the  way in which they are combined to form a Grid LSTM. 

\begin{figure}
\centering
\begin{minipage}{.5\textwidth}
\centering
\includegraphics[height=0.2\textheight]{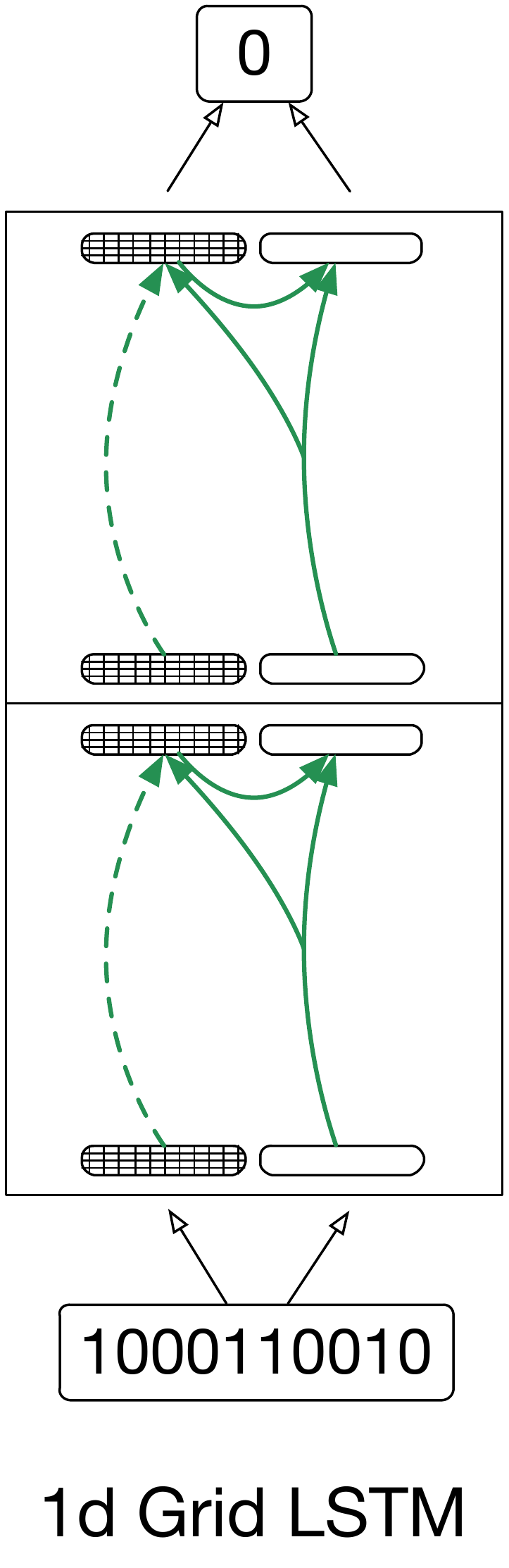}
\end{minipage}%
\begin{minipage}{.5\textwidth}
\centering
\includegraphics[height=0.2\textheight]{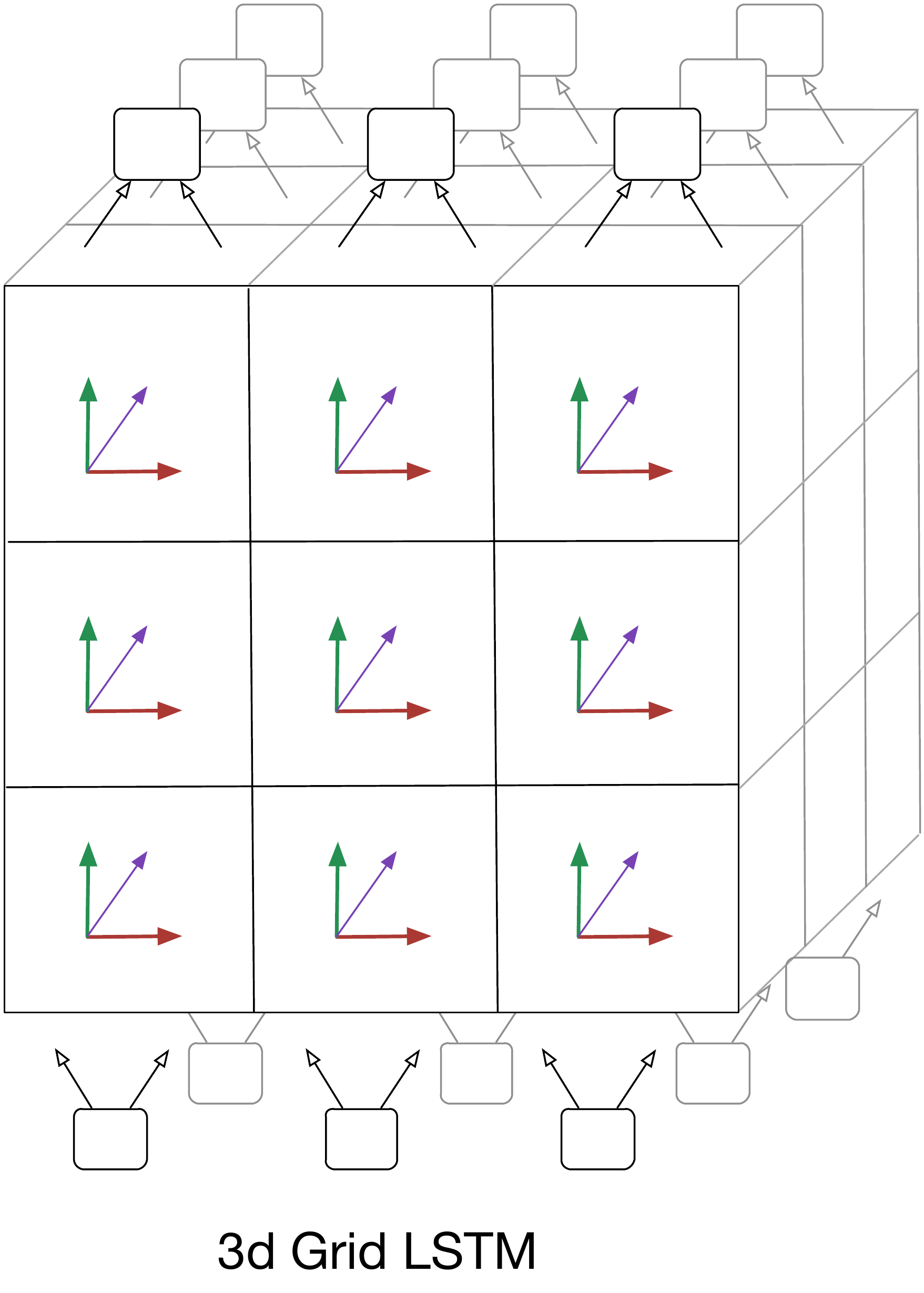}
\end{minipage}
\caption{Instances of one-dimensional and three-dimensional Grid LSTM. The network to the left is used for the parity results in the appendix. The translation and MNIST models below are specific instances of the 3d Grid LSTM to the right.}
\end{figure}

\subsection{Grid LSTM Blocks} 

As in multidimensional LSTM, a \emph{N}-dimensional block in a Grid LSTM receives as input $N$ hidden vectors $\vec{h}_1,...,\vec{h}_N$ and $N$ memory vectors $\vec{m}_1,...,\vec{m}_N$. Unlike the multidimensional case, the block outputs $N$ hidden vectors $\vec{h}_1',...,\vec{h}_N'$ and $N$ memory vectors $\vec{m}_1',...,\vec{m}_N'$ that are all distinct. 

The computation is simple and proceeds as follows. The model first concatenates the input hidden vectors from the \emph{N} dimensions:
\begin{equation}
\label{hh}
\vec{H} = \begin{bmatrix}
    \vec{h}_{1}    \\
    \vdots \\
    \vec{h}_{N}  \\
\end{bmatrix}
\end{equation}
Then the block computes $N$ transforms $\mathsf{LSTM}(\cdot, \cdot, \cdot)$, one for each dimension, obtaining the desired output hidden and memory vectors:
\begin{align}
\begin{split}
(\vec{h}_1', \vec{m}_1') &= \mathsf{LSTM}(\vec{H},\vec{m}_1,\vec{W}_1) \\
&\hspace{2mm} \vdots \\
(\vec{h}_N', \vec{m}_N') &= \mathsf{LSTM}(\vec{H},\vec{m}_N,\vec{W}_N) 
\end{split}
\end{align}
Each transform has distinct weight matrices $\vec{W}_i^u,\vec{W}_i^f,\vec{W}_i^o,\vec{W}_i^c$  in $\mathbb{R}^{d \times Nd}$ and applies the standard LSTM mechanism across the respective dimension. Note how the vector $\vec{H}$ that contains all the input hidden vectors is shared across the transforms, whereas the input memory vectors affect the $N$-way interaction but are not directly combined. \emph{N}-dimensional blocks can naturally be arranged in a \emph{N}-dimensional grid forming a Grid LSTM. As for a block, the grid has $N$ sides with \emph{incoming} hidden and memory vectors and $N$ sides with \emph{outgoing} hidden and memory vectors. Note that a block does not receive a separate data representation. A data point is projected into the network via a pair of input hidden and memory vectors along one of the sides of the grid.

\subsection{Priority Dimensions}

In a \emph{N}-dimensional block the transforms for all dimensions are computed in parallel. But it can be useful for a dimension to know the outputs of the transforms from the other dimensions, especially if the outgoing vectors from that dimension will be used to estimate the target. For instance, to prioritize the first dimension of the network, the block first computes the $N-1$ transforms for the other dimensions obtaining the output hidden vectors $\vec{h}_2',...,\vec{h}_N'$. Then the block concatenates these output hidden vectors  and the input hidden vector $\vec{h}_1$ for the first dimension into a new vector $\vec{H}'$ as follows:
\begin{equation}
\label{hh1}
\vec{H'} = \begin{bmatrix}
    \vec{h}_{1}    \\
    \vec{h}'_{2}    \\
    \vdots \\
    \vec{h}'_{N}  \\
\end{bmatrix}
\end{equation}
The vector is then used in the final transform to obtain the prioritized output hidden and memory vectors $\vec{h}_1'$ and $\vec{m}_1'$.

\subsection{Non-LSTM Dimensions}

In Grid LSTM networks that have only a few blocks along a given dimension in the grid, it can be useful to just have regular connections along that dimension without the use of cells. This can be naturally accomplished inside the block by using for that dimension in Eq.~6 a simple transformation with a nonlinear activation function instead of the transform $\mathsf{LSTM}(\cdot, \cdot, \cdot)$. Given a weight matrix $\vec{V} \in \mathbb{R}^{d\times Nd}$, for the first dimension this looks as follows:
\begin{align}
\begin{split}
\vec{h}_1' &= \alpha(\vec{V} * \vec{H}) \\
\end{split}
\end{align}
where $\alpha$ is a standard nonlinear transfer function or simply the identity. This allows us to see how, modulo the differences in the mechanism inside the blocks, Grid LSTM networks generalize the models in Sect.~2. A 2d Grid LSTM applied to temporal sequences with cells in the temporal dimension but not in the vertical depth dimension, corresponds to the Stacked LSTM. Likewise, the 3d Grid LSTM without cells along the depth corresponds to Multidimensional LSTM, stacked with one or more layers.

\subsection{Inputs from Multiple Sides}

If we picture a \emph{N}-dimensional block as in Fig.~1, we see that \emph{N} of the sides of the block have input vectors associated with them and the other \emph{N} sides have output vectors. As the blocks are arranged in a grid, this separation extends to the grid as a whole; each side of the grid has either input or output vectors associated with it. In certain tasks that have inputs of different types, a model can exploit this separation by projecting each type of input on a different side of the grid. The mechanism inside the blocks ensures that the hidden and memory vectors from the different sides will interact closely without being conflated. This is the case in the neural translation model introduced in Sect.~4 where source words and target words are projected on two different sides of a Grid LSTM.

\subsection{Weight Sharing}

Sharing of weight matrices can be specified along any dimension in a Grid LSTM and it can be useful to induce invariance in the computation along that dimension. As in the translation and image models, if multiple sides of a grid need to share weights, capacity can be added to the model by introducing into the grid a new dimension without sharing of weights. If the weights are shared along all dimensions including the depth,  we refer to the model as a \emph{Tied $N$-LSTM}.

\section{Experiments}

\subsection{Addition}

We first experiment with $2$-LSTM networks on learning to sum two 15-digit integers. The problem formulation is similar to that in \citep{DBLP:journals/corr/ZarembaS14}, where each number is given to the network one digit at a time and the result is also predicted one digit at a time. The input numbers are separated by delimiter symbols and an end-of-result symbol is predicted by the network; these symbols as well as input and target padding are indicated by $-$. An example is as follows: 
\begin{equation*}
\small
\begin{array}{ c c c c c c c c c c c c c c c  } 
- &1 & 2 & 3 & - & 8 & 9 & 9 & -  & - & - & - & -   \\
\\
 &  &  &  && &  \big \Downarrow \\
\\
- & - & - & - & - & - & - & - &  1 & 0 & 2 & 2 & - \\
\end{array}
\end{equation*}
Contrary to the work in \citep{DBLP:journals/corr/ZarembaS14} that uses from 4 to 9  digits for the input integers, we fix the number of digits to 15, we do not use curriculum learning strategies and we do not put digits from the partially predicted output back into the network, forcing the network to remember its partial predictions and making the task more challenging. The predicted output numbers  have either 15 or 16 digits.

\begin{figure}
\centering
\begin{minipage}{.48\textwidth}
\small
\begin{tabular}{ l | c | c | c   c c  } 
   &\!Layers\!&\!Samples\!&\!Accuracy\!\!\\ \hline \hline
 Stacked LSTM   & 1 & 5M & 51\% \\ \hline
  Untied 2-LSTM    & 5 & 5M & 67\% \\ \hline
  Tied 2-LSTM  & 18 & 0.55M &  $>99\%$  \\  \hline
\end{tabular}
\end{minipage}%
\begin{minipage}{.5\textwidth}
\centering
\includegraphics[width=0.8\textwidth]{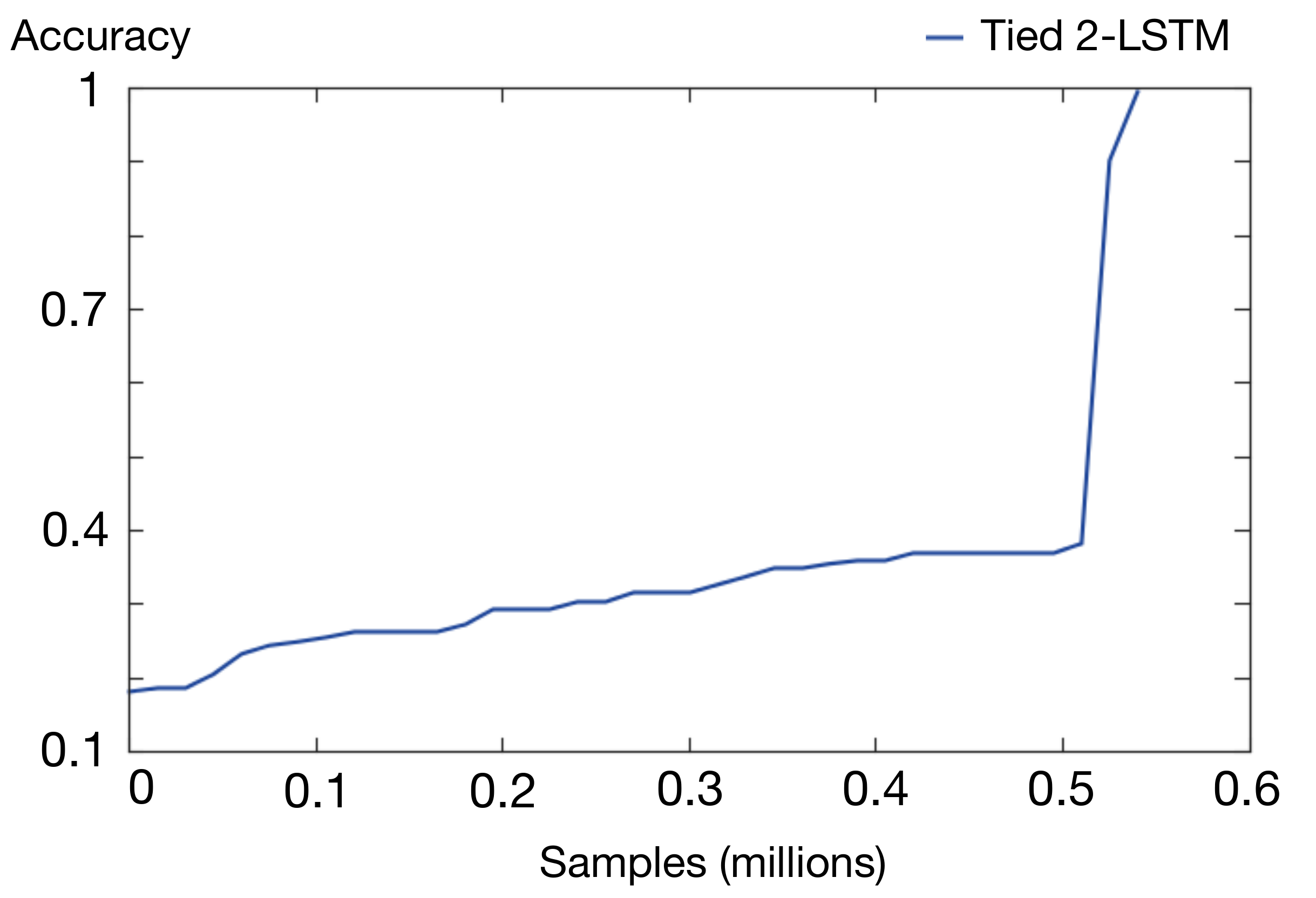}
\end{minipage}
\caption{Results on 15-digit addition. The left table gives results for the best performing networks of each type. The right graph depicts the learning curve of the 18-layer tied $2$-LSTM that solves the problem with less than 550K examples. The spike in the curve is likely due to the repetitions in the steps of the addition algorithm. }
\end{figure}

We compare the performance of $2$-LSTM networks with that of standard \emph{Stacked LSTM} (Fig.~2). We train the two types of networks with either tied or untied weights, with 400 hidden units each and with between 1 and 50 layers. We train the network with stochastic gradient descent using mini-batches of size 15 and the Adam optimizer with a learning rate of 0.001 \citep{DBLP:journals/corr/KingmaB14}. We train the networks for up to $5$ million samples or until they reach 100\% accuracy on a random sample of 100 unseen addition problems. Note that since during training all samples are randomly generated, samples are seen only once and it is not possible for the network to overfit on training data. The training and test accuracies agree closely.

Figure~4 relates the results of the experiments on the addition problem. The best performing tied $2$-LSTM is 18 layers deep and learns to perfectly solve the task in less than 550K training samples. We find that tied $2$-LSTM networks generally perform better than untied $2$-LSTM networks, which is likely due to the repetitive  nature of the steps involved in the addition algorithm. The best untied $2$-LSTM network has 5 layers, learns more slowly and achieves a per-digit accuracy of 67\% after 5 million examples. $2$-LSTM networks in turn perform  better than either tied or untied Stacked LSTM networks, where more stacked layers do not improve over the single-layer models. We see that the cells present a clear advantage for the deep $2$-LSTM networks by helping to mitigate the vanishing of gradients along the depth dimension.

\begin{figure}
\centering
\includegraphics[width=\textwidth]{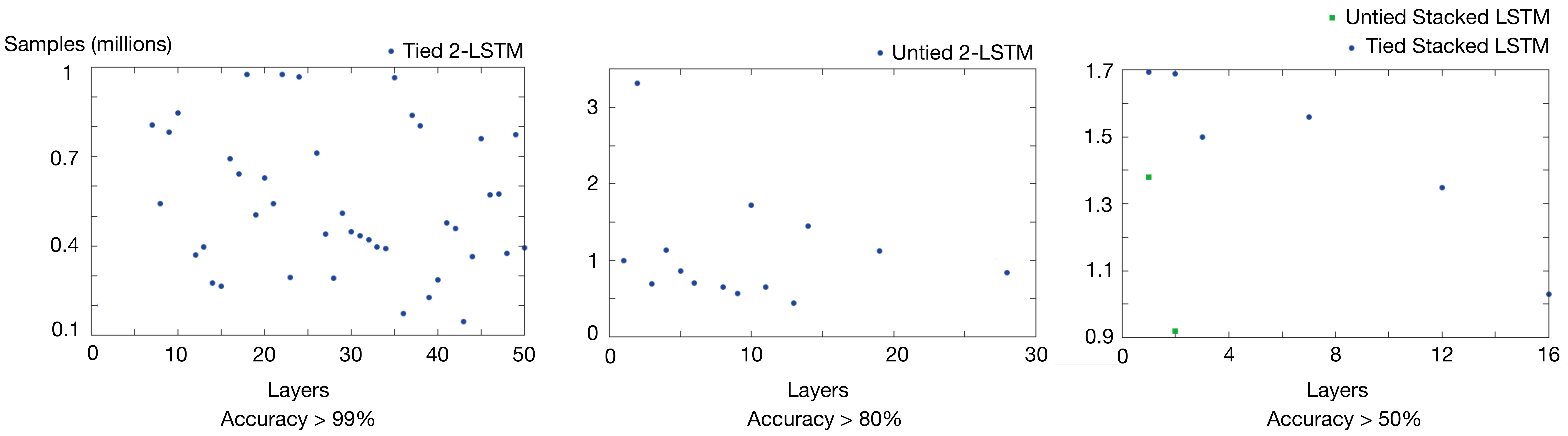}
\caption{Each dot in the three plots corresponds to a neural network of the respective type that has reached the accuracy of, respectively, $>99\%$, $>80\%$ and $>50\%$  at the memorization task. The networks all have 100 hidden units and the number of layers are indicated on the horizontal axis. The vertical axis indicates the number of samples needed to achieve the threshold accuracy. We see that deeper networks tend to learn faster than shallower ones, and that  $2$-LSTM networks are  more effective than Stacked LSTM networks in both the tied and untied settings. }
\end{figure}

\subsection{Memorization}

For our third algorithmic task, we analyze the performance of $2$-LSTM networks on the task of memorizing a random sequence of symbols. The sequences are 20 symbols long and we use a vocabulary of 64 symbols encoded as one-hot vectors and given to the network one symbol per step. The setup is similar to the one for addition above. The network is tasked with reading the input sequence and outputting the same sequence unchanged:
\begin{equation*}
\begin{array}{ c c c c c c c c c c c c c c c  } 
\small
- &\alpha & \beta & \gamma & - & - & - & -    \\
\\
 &  &  & \multicolumn{2}{c}{ \big \Downarrow }\\
\\
- &- & - & - &  \alpha & \beta & \gamma & - \\
\end{array}
\end{equation*}
Since the sequences are randomly generated, there is no correlation between successive symbols and  the network must memorize the whole sequence without compression.

We train $2$-LSTM and Stacked LSTM with either tied or untied weights on the memorization task. All  networks have 100 hidden units and have between 1 and 50 layers. We use mini-batches of size 15 and optimize the network using Adam  and a learning rate of 0.001. As above, we train each network for up to 5 million samples or until they reach 100\% accuracy on  100 unseen samples. Accuracy is measured per individual symbol, not per sequence. We do not use curriculum learning or other training strategies.

Figure~5 reports the performance of the networks. The small number of hidden units contributes to making the training of the networks difficult. But we see that tied $2$-LSTM networks are most successful and learn to solve the task with the smallest number of samples. The 43-layer tied $2$-LSTM network learns a solution with less than 150K samples. Although there is fairly high variance amid the solving networks, deeper networks tend to learn faster. In addition, there is large difference in the performance of tied $2$-LSTM networks and tied Stacked LSTM networks. The latter perform with much lower accuracy and Stacked LSTM networks with more than 16 layers do not reach an accuracy of more than 50\%. Here we see that the optimization property of the cells in the depth dimension delivers a large gain. Similarly to the case of the addition problem, both the untied $2$-LSTM networks and the untied Stacked LSTM networks take significantly longer to learn than the respective counterparts with tied weights, but the advantage of the cells in the depth direction clearly emerges for untied $2$-LSTM networks too.

\begin{figure}
\small
\begin{center}
\begin{tabular}{  l | c | c | c | c c  } 
  & BPC & Parameters & Alphabet Size & Test data \\ \hline \hline

 Stacked LSTM  \citep{DBLP:journals/corr/Graves13} & 1.67 & 27M & 205 & last 4MB \\ \hline
    MRNN \citep{sutskever11rnn}    & 1.60 & 4.9M & 86  & last 10MB \\ \hline
  GFRNN \citep{DBLP:journals/corr/ChungGCB15} & 1.58  &  20M  & 205 & last 5MB \\ \hline
  \textbf{Tied 2-LSTM} & \textbf{1.47} & 16.8M & 205 & last 5MB \\ \hline 

\end{tabular}
\end{center}
\caption{Bits-per-character results for various models measured on the Wikipedia dataset together with the respective number of parameters and the size of the alphabet that was used.  Note the slight differences in test data and alphabet size.}
\end{figure}

\subsection{Character-Level Language Modelling}

We next test the $2$-LSTM network on the Hutter challenge Wikipedia dataset \citep{hutterDataset}. The aim is to successively predict the next character in the corpus. The dataset has 100 million characters. We follow the splitting procedure of \citep{DBLP:journals/corr/ChungGCB15}, where the last 5 million characters are used for testing. The alphabet has 205 characters in total. 

We use a tied $2$-LSTM with 1000 hidden units and 6 layers of depth. As in Fig.~2 and in the previous tasks, the characters are projected both to form the initial input hidden and cell vectors and the top softmax layer is connected to the topmost output hidden and cell vectors. The model has a total of $2000 \times 4000 + 205 \times 4 \times 1000 = 8.82 \times 10^6$ parameters. As  usual the objective is to minimize the negative log-likelihood of the character sequence under the model. Training is performed by sampling sequences of 10000 characters and processing them in order. We back propagate the errors every 50 characters. The initial cell and hidden vectors in the temporal direction are initialized to zero only at the beginning of each sequence; they maintain their forward propagated values after each update in order to simulate full back propagation. We use mini-batches of 100, thereby processing 100 sequences of 10000 characters each in parallel. The network is trained with Adam with a learning rate of 0.001 and training proceeds for approximately 20 epochs. 

Figure~6 reports the bits-per-character performance together with the number of parameters of various recently proposed models on the dataset. The tied $2$-LSTM significantly outperforms other models despite having fewer parameters. More layers of depth and adding capacity by untying some of the weights are likely to further enhance the $2$-LSTM.

\subsection{Translation}

We next use the flexibility of Grid LSTM to define a novel neural translation model. In the neural approach to machine translation one trains a neural network end-to-end to map the source sentence to the target sentence \citep{kalchbrenner13emnlp,sutskever2014sequence,DBLP:journals/corr/ChoMGBSB14}. The mapping is usually performed within the \emph{encoder-decoder} framework. A neural network, that can be convolutional or recurrent, first encodes the source sentence and then the computed representation of the source conditions a recurrent neural network to generate the target sentence. This approach has yielded strong empirical results, but it can suffer from a bottleneck. The encoding of the source sentence must contain information about all the words and their order; the decoder network in turn cannot easily revisit the unencoded source sentence to make decisions based on partially produced translations. This issue can be alleviated by a soft attention mechanism in the decoder neural network that uses gates to focus on specific parts of the source sentence \citep{DBLP:journals/corr/BahdanauCB14}.

\begin{figure}
\centering
\includegraphics[width=0.5\textwidth]{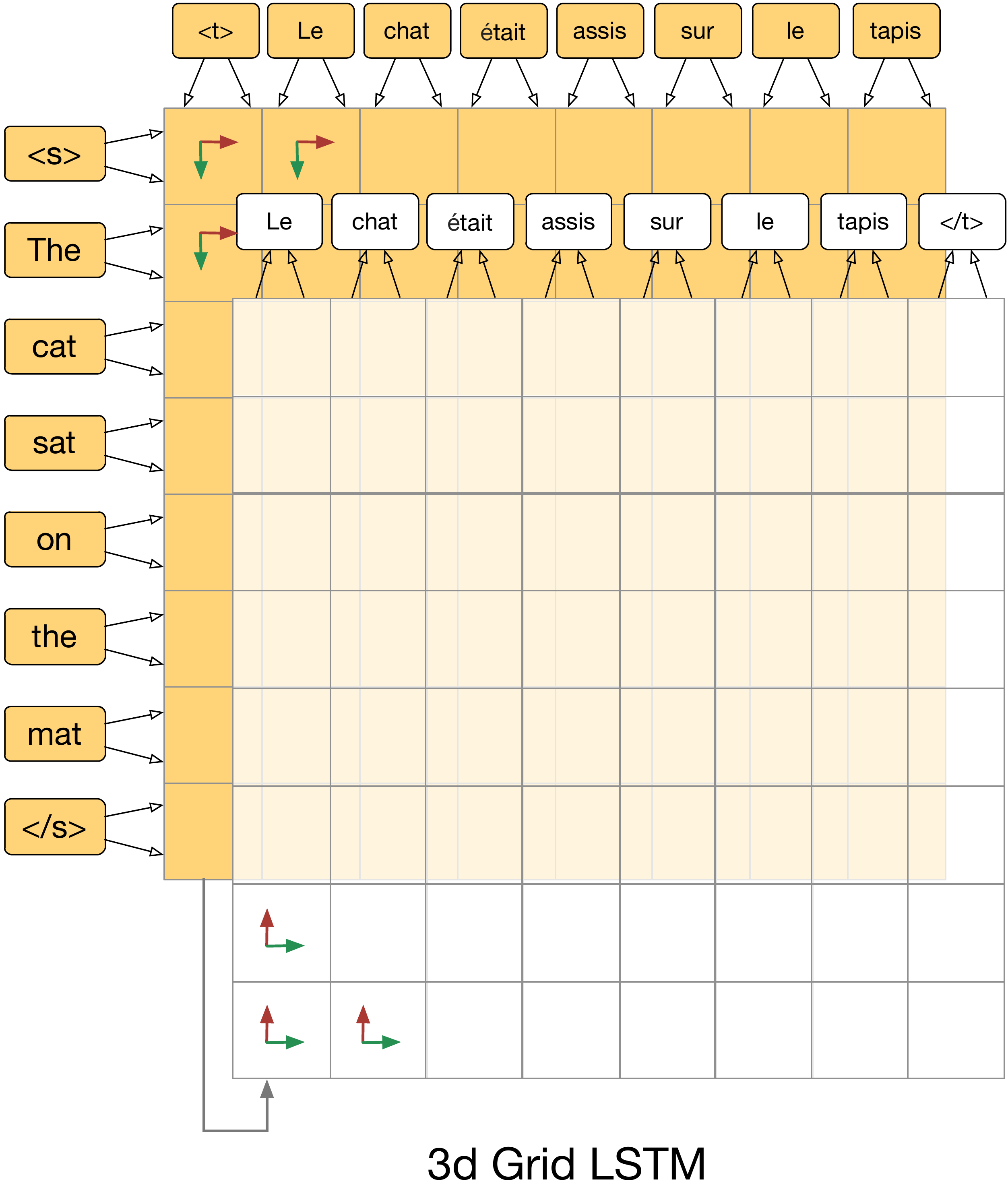}
\caption{Illustration of the $3$-LSTM neural translation model. }
\end{figure}

We use Grid LSTM to view translation in a novel fashion as a \emph{two-dimensional} mapping. We call this the \emph{Reencoder} network. One dimension processes the source sentence whereas the other dimension produces the target sentence. The resulting network repeatedly \emph{re-encodes} the source sentence conditioned on the part of the target sentence generated so far, thus functioning as an implicit attention mechanism. The size of the representation of the source sentence varies with length and the source sentence is repeatedly scanned based on each generated target word. As represented in Fig.~9, for each target word, beginning with the start-of-target-sentence symbol, the network scans the source sentence  one way in the first layer and the other way in the second layer; the scan depends on all the target words that have been generated so far and at each block the two layers communicate directly. Note that, like the attention-based model \citep{DBLP:journals/corr/BahdanauCB14}, the two-dimensional translation model has complexity $O(nm)$, where $n$ and $m$ are respectively the length of the source and target; by contrast the recurrent encoder-decoder model only has complexity $O(m+n)$. This  gives additional computational capacity to the former models.

 Besides addressing the bottleneck, the two-dimensional setup aims at explicitly capturing the invariance present in translation. Translation patterns between two languages are invariant above all to \emph{position} and \emph{scale} of the pattern.  For instance, reordering patterns - such as the one that maps the English \emph{``do not $\langle$verb$\rangle$"} to the French \emph{``ne  $\langle$verb$\rangle$ pas"}, or the one that sends a part of an English verb to the end of a German sentence -  should be detected and applied independently of where they occur in the source sentence or of the number of words involved in that instance of the pattern. To capture this, the Grid LSTM translation model shares the weights across the source and target dimensions. In addition, a hierarchy of stacked two-dimensional grids in opposite directions is used to both increase capacity and help with learning longer scale translation patterns. The resulting model is a three-dimensional Grid LSTM where hierarchy grows along the third dimension. The model is depicted in Fig.~7.

\begin{figure}
\centering
\small
\begin{tabular}{  l | c | c | c |  c  c  c c  } 
  & Valid-1 & Test-1 & Valid-15 & Test-15 \\ \hline \hline
DGLSTM-Attention \citep{DBLP:journals/corr/YaoCVDD15} & - & 34.5 & - & - \\ \hline
CDEC  \citep{cdec} & 30.1 & 41 & 50.1 & 58.9 \\ \hline
\textbf{3-LSTM} (7 Models) & {30.3} & {42.4} & {51.8} & {60.2} \\ \hline
\end{tabular}
\centering
\tiny
\begin{center}
\vspace{0.5cm}
\small
\begin{tabular}{  c | c  c  c  c c  } \hline
 Reference &  thank you . please pay for this bill at the cashier . \\ 
 Generated & thank you , ma 'am . please give this bill to the cashier and pay there . \\ \hline \hline
Reference & how about having lunch with me some day ? i found a good restaurant near my hotel . \\ 
Generated & how about lunch with me ? i found a good restaurant near my hotel . \\ \hline
\end{tabular}
\end{center}
\caption{The first table contains BLEU-4 scores of the $3$-LSTM neural translation model, the CDEC system and the Depth-Gated LSTM (DGLSTM) with attention mechanism; the scores are  calculated against either the main reference translation or against the 15 available reference translations in the BTEC corpus. CDEC is a state-of-the-art hierarchical phrase based system with many component models. The second table contains examples of generated translations.}
\end{figure}

We evaluate the Grid LSTM translation model on the IWSLT BTEC Chinese-to-English corpus that consists of 44016 pairs of source and target sentences for training, 1006 for development and 503 for testing. The corpus has about 0.5M words in each language, a source vocabulary of 7055 Chinese words and a target vocabulary of 5646 English words (after replacing words that occur only once with the UNK symbol). Target sentences are on average around 12 words long. The development and test corpora come with 15 reference translations. The $3$-LSTM uses two two-dimensional grids of 3-LSTM blocks for the hierarchy. Since the network has just two layers in the third dimension, we use regular identity connections without nonlinear transfer function along the third dimension, as defined in Sect.~3.3; the source and target dimensions have tied weights and LSTM cells. The processing is bidirectional, in that the first grid processes the source sentence from beginning to end and the second one from end to beginning. This allows for the shortest distance that the signal travels between input and output target words to be constant and independent of the length of the source. Note that the second grid receives an input coming from the grid below at each $3$-LSTM block. We train seven models with vectors of size 450 and apply dropout with probability 0.5 to the hidden vectors within the blocks. For the optimization we use Adam with a learning rate of 0.001. At decoding the output probabilities are averaged across the models. The beam search has size 20 and we discard all candidates that are shorter than half of the length of the source sentence.
The results are shown in Fig.~8. Our best model reaches a perplexity of 4.54 on the test data. We use as baseline the state-of-the-art hierarchical phrase-based system CDEC \citep{cdec}. We see that the Grid LSTM significantly outperforms the baseline system on both the validation and test data sets.

\section{Conclusion}

We have introduced Grid LSTM, a network that uses LSTM cells along all of the dimensions and modulates in a  novel fashion the multi-way interaction. We have seen the advantages of the cells compared to regular connections in solving tasks such as parity, addition and memorization. We have described powerful and flexible ways of applying the model to character prediction, machine translation and image classification, showing strong performance across the board. 

\subsection*{Acknowledgements}

We thank  Koray Kavukcuoglu, Razvan Pascanu, Ilya Sutskever and Oriol Vinyals for helpful comments and discussions.

\begingroup

\small{
\bibliographystyle{iclr2016_conference}
\bibliography{DualLSTM}
}
\endgroup

\section*{Appendix}

We here report on two additional results, one algorithmic and the other one empirical, where we see that without special initialization or training tricks, a 1-LSTM network can learn to compute parity for up to 250 input bits, and a 3-LSTM network applied to images obtains strong results on MNIST.

\subsection{Parity}

We apply one-dimensional Grid LSTM to learning parity. Given a string $b_1,...,b_k$ of $k$ bits 0 or 1, the \emph{parity} or \emph{generalized XOR} of the string is defined to be 1 if the sum of the bits is odd, and 0 if the sum of the bits is even. Although manually crafted neural networks for the problem have been devised \citep{DBLP:journals/nn/HohilLS99}, training a generic neural network from a finite number of examples and a generic random initialization of the weights to successfully learn to compute the parity of $k$-bit strings for significant values of $k$ is a longstanding problem \citep{minskypapert,duch06}. It is core to the problem that the $k$-bit string is given to the neural network as a whole through a single projection; considering one bit at a time and remembering the previous partial result in a recurrent or multi-step architecture reduces the problem of learning $k$-bit parity to the simple one of learning just $2$-bit parity. Learning parity is difficult because a change in a single  bit in the input changes the target value and the decision boundaries in the resulting space are highly non-linear.

We train 1-LSTM networks with tied weights and we compare them with fully-connected feed-forward networks with ReLU or $\tanh$ activation functions and with either tied or untied weights. We search the space of hyper-parameters  as follows. The 1-LSTM networks are trained with either 500 or 1500 hidden units and having from 1 to 150 hidden layers. The 1-LSTM networks are trained on input strings that have from $k=20$ to $k=250$ bits in increments of 10. The feed-forward ReLU and $\tanh$ networks are trained with 500, 1500 or 3000 units and also having from 1 to 150 hidden layers. The latter networks are trained on input bit strings that have between $k=20$ and $k=60$ bits in increments of 5. Each network is trained with a maximum of $10$ million samples or four days of computation on a Tesla K40m GPU. For the optimization we use mini-batches of size 20 and the AdaGrad rule with a learning rate of 0.06 \citep{Duchi:EECS-2010-24}. A network is considered to have found the solution if the network correctly computes the parity of 100 randomly sampled unseen $k$-bit strings. Due to the nature of the problem, during training the predicted accuracy is never better than random guessing and when the network finds a solution the accuracy suddenly spikes to 100\%.

\begin{figure}
\label{parityres}
\centering
\includegraphics[width=0.95\textwidth]{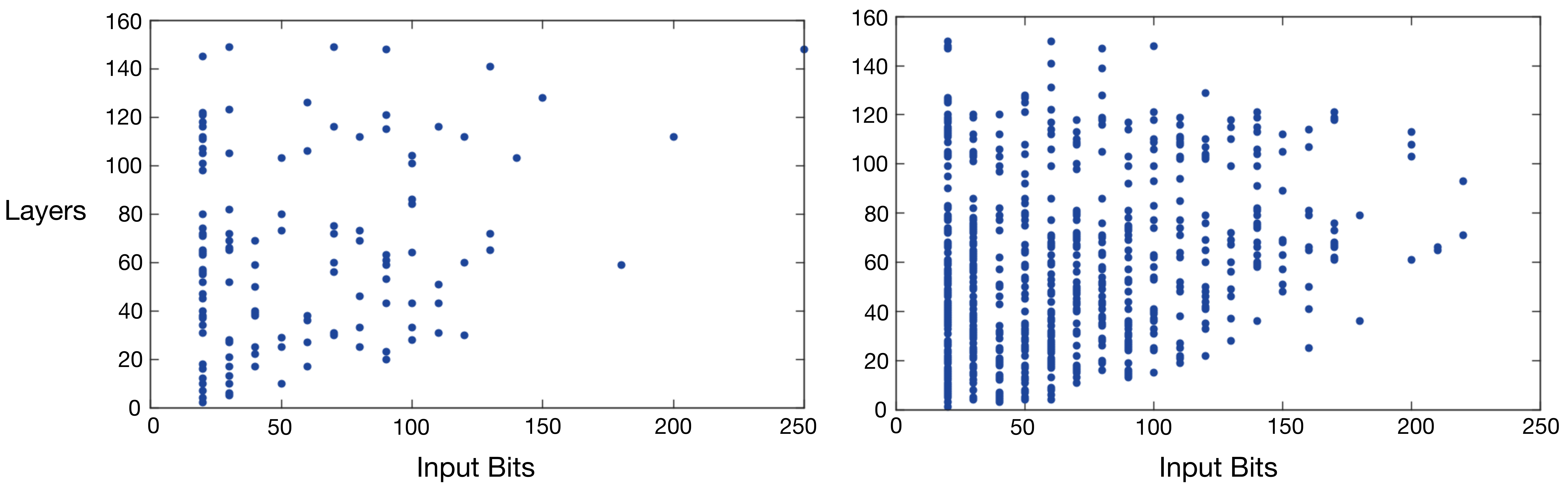}
\caption{Results on training tied 1-LSTM networks to compute the $k$-bit parity of $k$ input bits. The left diagram contains solutions found with 1-LSTM networks with 500 hidden units, whereas the right diagram shows solutions found with  1-LSTM networks with 1500 units. The horizontal axis corresponds to the number $k$ of input bits. The vertical axis corresponds to the number of  layers in the networks. Each point in the diagram corresponds to $100\%$ classification accuracy of the respective network on a sample of 100 unseen $k$-bit strings. The networks see up to 10 million bit strings during training but often find solutions with many fewer strings. Missing points in the diagram indicate failure to find a solution within the training set size or time constraints.  }
\end{figure}
\begin{figure}
\begin{minipage}{.52\textwidth}
\centering

\begin{tabular}{ l | c | c | c   c c  } 
   &\!Layers\!&\!Hidden\!&\! Input Bits $k$\!\!\\ \hline \hline
 Tied Tanh FFN   & 5 & 1500 & 30 \\ \hline
  Tied ReLU FFN    & 4 & 1500 & 30 \\ \hline
  Tied 1-LSTM    & 72 & 1500 & 220 \\ \hline 
    Tied 1-LSTM    & 148 & 500 & 250 \\ \hline
\end{tabular}
\end{minipage}
\begin{minipage}{.5\textwidth}
\centering
\includegraphics[width=0.9\textwidth]{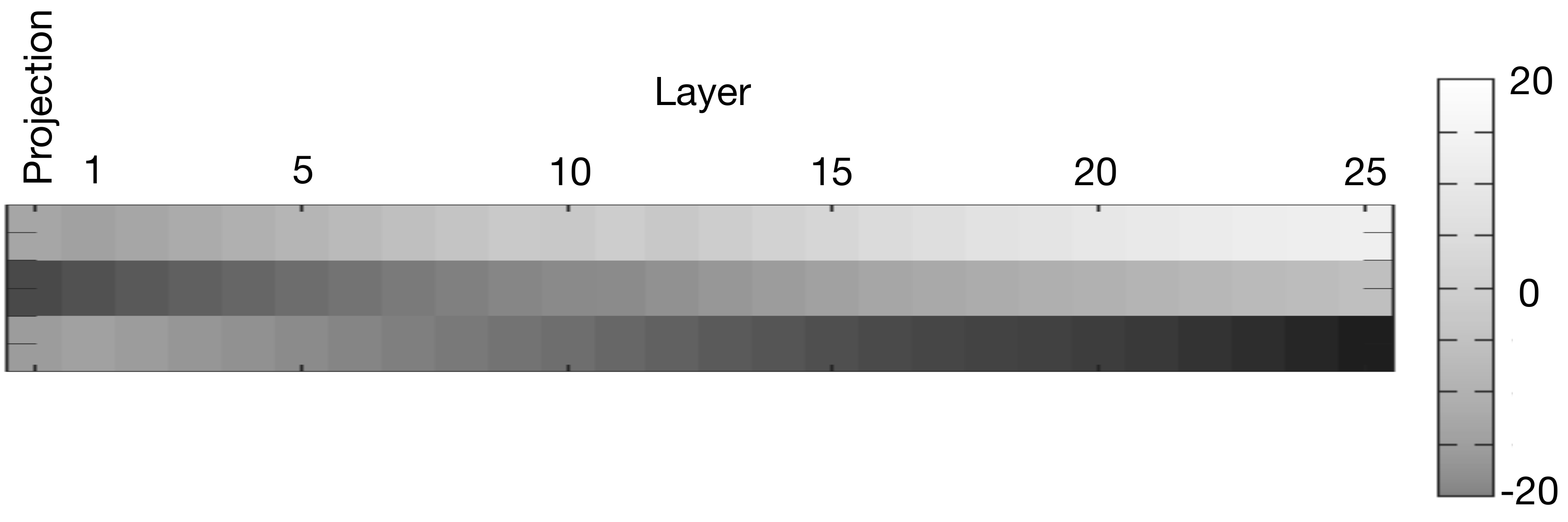}
\end{minipage}
\caption{The left table reports the best performing networks on $k$-bit parity. The right figure is a heat map of activation values of selected \emph{counter neurons} in a 1-LSTM network that has 25 layers and is trained on the parity of 50-bit strings. The specific values are obtained by a feed-forward pass through the network using as input the bit string $0^{10}1^{40}$; different bit strings gave similar results. }
\end{figure}

Figure~9 depicts the results of the experiments with $1$-LSTM networks and Figure~10 relates the best performing networks of each type. For the feed-forward ReLU and $\tanh$ networks with either tied or untied weights, we find that these networks fail to find solutions for $k=35$ bits and beyond.  Some networks in the search space find solutions for $k=30$ input bits. By contrast, as represented in Fig.~9, tied $1$-LSTM networks find solutions for up to $k=250$ bits.  

There appears to be a correlation between the length $k$ of the input bit strings and the minimum depth of the $1$-LSTM networks. The minimum depth of the networks increases with $k$ suggesting that longer bit strings need more operations to be applied to them; however, the rate of growth is \emph{sub-linear} suggesting that more than a single bit of the input is considered at every step. We visualized the activations of the memory vectors obtained via a feed-forward pass through one of the $1$-LSTM networks using selected input bit strings (Fig.~10). This revealed the prominent presence of \emph{counting neurons} that keep a counter for the number of layers processed so far. These two aspects seem to suggest that the networks are using the  cells to process the bit string \emph{sequentially} by attending to parts of it at each step in the computation, a seemingly crucial feature that is not available in ReLU or $\tanh$ transfer functions.

\subsection{MNIST Digit Recognition}

\begin{figure}
\centering
\includegraphics[width=0.7\textwidth]{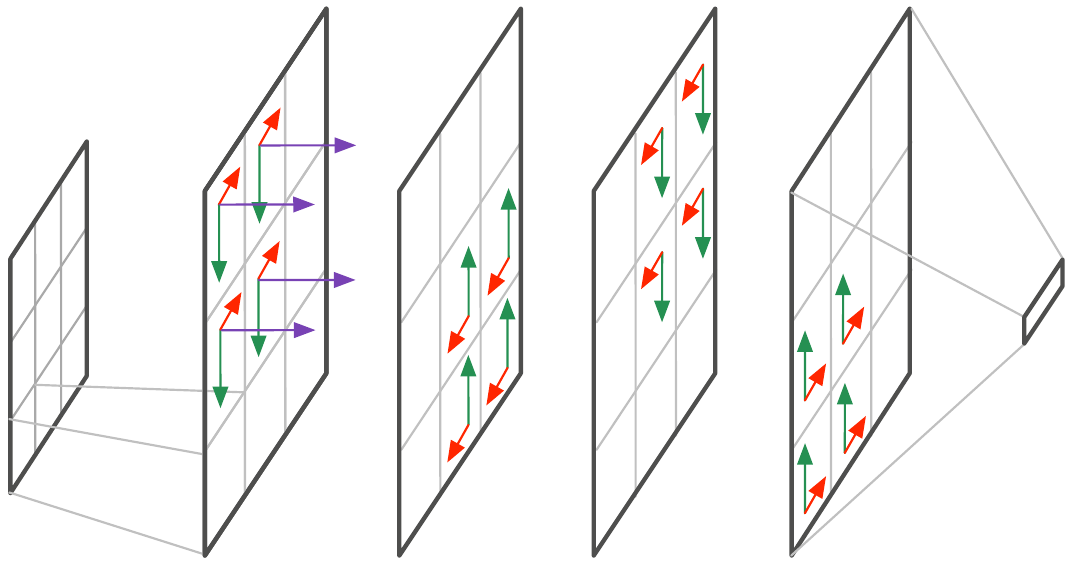}
\caption{A 3-LSTM network applied to non-overlapping patches of an image. Each patch is projected to form the input hidden and cell vectors of the depth dimension of the 3-LSTM blocks. The arrows across the spatial dimensions indicate the flow of the computation for that layer. No subsampling or pooling occurs in the networks as the topmost layer simply concatenates all the output hidden and memory vectors of the depth dimension, passes them through a layer of ReLUs and the final softmax layer.  }
\end{figure}

In our last experiment we apply a $3$-LSTM network to images. We consider non-overlapping patches of pixels in an image as forming a two-dimensional grid of inputs. The $3$-LSTM performs computations with LSTM cells along three different dimensions. Two of the dimensions correspond to the two spatial dimensions of the grid, whereas the remaining dimension is the depth of the network. Like in a convolutional neural network \citep{lecun98convolution}, the same three-way transform of the $3$-LSTM is applied at all parts of the grid, ensuring that the same features can be extracted across all parts of the input image. Due to the unbounded context size of the $3$-LSTM, the computations of features at one end of the image can be influenced by the features computed at the other end of the image within the same layer. Due to the cells along the depth direction, features from the present patch can be passed onto the next layer either unprocessed or as processed by the layer itself as a function of neighboring patches. 

We construct the network as depicted in Fig.~11. We divide the $28  \times 28$ MNIST image into $p \times p$ pixel patches, where $p$ is a small number such as 2 or 4. The patches are then \emph{linearized} and projected into two vectors of the size of the hidden layer of the $3$-LSTM; the projected vectors are the input hidden and memory vectors at the first layer in the depth direction of the $3$-LSTM. At each layer the computation of the $3$-LSTM starts from one corner of the image, follows the two spatial dimensions and ends in the opposite corner of the image. The network has a few layers of depth, each layer starting the computation at one of the corners of the image. In the current form there is no pooling between successive layers of the $3$-LSTM. The topmost layer concatenates all the output hidden and memory vectors at all parts of the grid. These are then passed through a layer of ReLUs and a final softmax layer.

The setup has some similarity with the original application of Multidimensional LSTM to images \citep{graves12supervised} and with the recently described ReNet architecture \citep{DBLP:journals/corr/VisinKCMCB15}. 
The difference with the former is that we apply multiple layers of depth to the image, use three-dimensional blocks and concatenate the top output vectors before classification. The difference with the ReNet architecture is that the $3$-LSTM processes the image according to the two inherent spatial dimensions; instead of stacking hidden layers as in the ReNet, the block also modulates directly what information is passed along the depth dimension.

The training details are as follows. The MNIST dataset consists of 50000 training images, 10000 validation images and 10000 test images. The pixel values are normalized by dividing them by 255. Data augmentation is performed by shifting training images from 0 to 4 pixels in the horizontal and vertical directions and padding with zero values. The shift in the two directions is chosen uniformly at random. Validation samples are used for retraining the best model settings found during the grid search. We train the $3$-LSTM both with and without cells in the depth dimension. The $3$-LSTM with the cells uses patches of $2 \times 2$ pixels, has four LSTM layers with 100 hidden units and one ReLU layer with 4096 units. The $3$-LSTM without the cells in the depth dimension has input patches of size $3\times 3$ obtained by cropping the image to a size of $27 \times 27$, it also has four LSTM layers of 100 units and has a ReLU layer of 2048 units.  For the latter model we use ReLU as transfer function for the depth direction as in Eq.~6. We use mini-batches of size 128 and train the models using Adam and a learning rate of 0.001.

Figure 12 reports test set errors of our models and that of competing approaches. We can see that even in the absence of pooling the $3$-LSTM with the cells performs near the state-of-the-art. The $3$-LSTM without the cells also performs quite well; the  cells in the depth direction likely help with the feature extraction at the higher layers. The other approaches, with the exception of ReNet, are convolutional neural networks.

\begin{figure}
\begin{center}
\begin{tabular}{ l | c c }
  & Test Error (\%)  \\ \hline \hline
Wan \emph{et al.} \citep{conf/icml/WanZZLF13} & {0.28}  \\  
Graham \citep{DBLP:journals/corr/Graham14} & {0.31}  \\  
 \textbf{Untied 3-LSTM} & \textbf{0.32}  \\  

Ciresan \emph{et al.} \citep{Ciresan:2012e} & {0.35}  \\  
  \textbf{Untied 3-LSTM with ReLU (*)} & \textbf{0.36}  \\  
 Mairar \emph{et al.} \citep{DBLP:journals/corr/MairalKHS14} & {0.39}  \\  
  Lee \emph{et al.} \citep{DBLP:conf/aistats/LeeXGZT15} & {0.39}  \\  
    Simard \emph{et al.} \citep{DBLP:conf/icdar/SimardSP03} & {0.4}  \\  
    Graham \citep{DBLP:journals/corr/Graham14a} & {0.44}  \\  

Goodfellow \emph{et al.} \citep{DBLP:conf/icml/GoodfellowWMCB13} & {0.45}  \\  
Visin \emph{et al.} \citep{DBLP:journals/corr/VisinKCMCB15} & {0.45}  \\  
  Lin \emph{et al.} \citep{DBLP:journals/corr/LinCY13} & {0.47}  \\  \hline
\end{tabular}

\end{center}
\caption{Test error on the MNIST dataset. All approaches are convolutional networks except for Visin \emph{et al.} that uses a stack of single-direction recurrent neural networks. (*) This Grid LSTM has non-LSTM connections along the depth only and uses the ReLU instead. }
\end{figure}

\end{document}